\pdfoutput=1

\documentclass[11pt]{article}

\usepackage{acl}

\usepackage{times}
\usepackage{latexsym}

\usepackage{comment}

\usepackage{microtype}
\usepackage{amsmath,amssymb}
\usepackage{comment}
\usepackage{adjustbox}
	
\usepackage{color, colortbl}

\usepackage{xspace}
\usepackage{booktabs}
\usepackage{multicol}
\usepackage{amsmath}
\usepackage{multirow}
\usepackage{graphicx}
\usepackage{enumitem}
\usepackage{mdwlist}
\usepackage{url}

\usepackage[T1]{fontenc}

\usepackage[utf8]{inputenc}

\usepackage{microtype}
\usepackage{url}

\definecolor{lightgray}{gray}{0.9}
\title{SemEval 2023 Task 9: Multilingual Tweet Intimacy Analysis}

\author{
  Jiaxin Pei $^{\clubsuit}$  ~ Vítor Silva$^{\dagger}$ ~ Maarten Bos $^{\dagger}$ ~ Yozen Liu$^{\dagger}$ \\
 \textbf{Leonardo Neves}$^{\dagger}$ ~ \textbf{David Jurgens}$^{\clubsuit}$ ~ \textbf{Francesco Barbieri}$^{\dagger}$ \vspace{0.1cm} \\
$^\clubsuit$School of Information, University of Michigan, Ann Arbor, MI, USA \vspace{0.1cm} \\
$^\dagger$Snap Inc, Santa Monica, CA, USA\\
{ \tt $^{\clubsuit}$\{pedropei, jurgens\}@umich.edu}\\
{  \tt $^{\dagger}$\{fbarbieri, vsilvasousa, maarten, yliu2, lneves\}@snap.com}\\
}

\newcommand\DATASET{\textsc{Mint}}

\begin{document}
\maketitle

\begin{abstract}
We propose \DATASET{}, a new \textbf{M}ultilingual \textbf{int}imacy analysis dataset covering 13,372 tweets in 10 languages including English, French, Spanish, Italian, Portuguese, Korean, Dutch, Chinese, Hindi, and Arabic. We benchmarked a list of popular multilingual pre-trained language models. The dataset is released along with the \href{https://sites.google.com/umich.edu/semeval-2023-tweet-intimacy}{SemEval 2023 Task 9: Multilingual Tweet Intimacy Analysis}. 
\end{abstract}

\section{Introduction}
Intimacy has long been viewed as a primary dimension of human relationships and interpersonal interactions \citep{maslow1981motivation,sullivan2013interpersonal, prager1995psychology}. Existing studies suggest that intimacy is an essential component of language and can be modeled with computational methods \citep{pei2020quantifying}. Textual intimacy is an important social aspect of language, and automatically analyzing it can help to reveal important social norms in various contexts \citep{pei2020quantifying}. Recognizing intimacy can also serve as an important benchmark to test the ability of computational models to understand social information \citep{hovy2021importance}. 

Despite the importance of intimacy in language, resources on textual intimacy analysis remain rare. \citet{pei2020quantifying} annotated the first textual intimacy dataset containing 2,397 English questions collected mostly from social media posts and fictional dialogues. However, phrases following the question structure are used for interrogative situations, and models trained over it may not generalize well over texts in other forms of languages.  

To further promote computational modeling of textual intimacy, we annotated a new multilingual textual intimacy dataset named \DATASET{}. \DATASET{} covers tweets in 6 languages as the training data, including English, Spanish, French, Portuguese, Italian, and Chinese, covering major languages used in The Americas, Europe, and Asia. A total of 12,000 tweets are annotated for the 6 languages. To test the model generalizability under the zero-shot settings, we also annotated small test sets for Dutch, Korean, Hindi, and Arabic (500 tweets for each). 

We benchmarked a series of large multilingual pre-trained language models including XLM-T \citep{barbieri2021xlm}, XLM-R \citep{conneau2019unsupervised}, BERT \citep{devlin2018bert}, DistillBERT \citep{Sanh2019DistilBERTAD} and MiniLM \citep{wang2020minilm}. We found that distilled models generally perform worse than other normal models, while the XLM-R model trained over the twitter dataset (XLM-T) performs the best on 7 languages. While the pre-trained language models are able to achieve promising performance, zero-shot prediction of unseen languages remains challenging especially for Korean and Hindi.

\section{Data}
We choose Twitter as the source of our dataset as Twitter is a public media platform that naturally includes multilingual text data, and from our analysis, a fair amount of intimate texts. In this section, we introduce the data collection and annotation process for \DATASET{}.

\subsection{Sampling}
We use tweets sampled from 2018 to 2022. We use the \textit{lang\_id} key in the tweet object to select English and Chinese tweets. For other languages, we use fastText \citep{joulin2016bag,joulin2016fasttext} for language identification\footnote{\url{https://fasttext.cc/docs/en/language-identification.html}} and assign language labels when the model confidence is larger than 0.8. All the mentions of unverified users are replaced with a special token ``@user'' during pre-processing to remove noise from random and very infrequent usernames.

We fine-tune XLM-T, a multilingual RoBERTa model adapted to the Twitter domain \citep{barbieri2021xlm} over the annotated question intimacy dataset \citep{pei2020quantifying}.
The fine-tuned model attained a Pearson's $r$ of 0.80 for English questions\footnote{\citet{pei2020quantifying} report a Pearson's $r$ of 0.82 using RoBERTa-base. Given that XLM-T is pre-trained on tweets, a Pearson's $r$ of 0.8 is reasonable.} and we use it to estimate the intimacy of all the collected tweets in 10 languages.  
Then, in the second step, we split the tweets into 5 buckets based on the estimated intimacy and up-sampled relatively more intimate tweets. We did bucket sampling for 1,000 English tweets and randomly sampled another 1,000 English tweets as well as all the tweets for the rest of the languages\footnote{We intended to do bucket sampling for all the data, however, due to an issue in the pre-processing, we were only able to do it for 1,000 tweets. We conducted further analyses for the potential effect of this error. We found that the distribution of the final annotated intimacy scores are not changed much while the fine-tuned XLM-T only achieved a Pearson's $r$ of 0.43 on the random sample, suggesting that the model trained on Reddit questions may not be reliable enough to detect intimacy in tweets. }.

\begin{table*}[t!]
\small
\newcommand{\tabincell}[2]{\begin{tabular}{@{}#1@{}}#2\end{tabular}}
\begin{center}
 \resizebox{0.99\textwidth}{!}{
\begin{tabular}{l|l}
\toprule
\textbf{English} & \textbf{Intimacy} \\
\midrule

Ukrainian Railways Chief Says `Honest' Belarusians Are Cutting Russian Supplies By Train http  &  1.00 \\ \\[-1em] \rowcolor{lightgray}
19:04h Temp: 28.9°F Dew Point: 19.40°F Wind:SSW 4.3mph Rain:0.00in. Baro:29.66 inHg via MeteoBridge 3.2  &  1.00 \\ \\[-1em]
A team that shops together stays together...helping life go right @StateFarm http  &  1.00 \\ \rowcolor{lightgray} & \\[-1em]
\rowcolor{lightgray}
Leicester City fans - keep an eye on Ross Barkley. Could be moving to the Foxes on a permanent for £11m... \#lcfc  &  1.25 \\ \\[-1em]
@user They aren’t open  &  1.25 \\ \rowcolor{lightgray} & \\[-1em]
\rowcolor{lightgray}
Kenya I vote for \#Butter for \#BestMusicVideo at the 2022 \#iHeartAwards @BTS\_twt  &  1.25 \\ \\[-1em]
That might have been the best episode of power ever  &  1.40 \\ \rowcolor{lightgray} & \\[-1em]
\rowcolor{lightgray}
@user Coming to USA if Trump loses in 2020.  &  1.40 \\ \\[-1em]
Change the formula to get a different result  &  1.60 \\ \rowcolor{lightgray} & \\[-1em]
\rowcolor{lightgray}
@user thank u  &  2.50 \\ \\[-1em]
@user Happy birthday!  &  2.60 \\ \rowcolor{lightgray} & \\[-1em]
\rowcolor{lightgray}
it’s the worst feeling when you feel like no matter how much u do for a person you’ll never get the same in return  &  3.00 \\ \\[-1em]
@user you’re not my mom &  3.00 \\ \rowcolor{lightgray} & \\[-1em]
\rowcolor{lightgray}
@user @user Love you  &  4.00 \\ \\[-1em]
I am SO ecstatic I’m not married to a man who has cheated on me.  &  4.33 \\ \rowcolor{lightgray} & \\[-1em]
\rowcolor{lightgray}
My nails so mf ghetto. I’m embarrassed &  4.67 \\ \\[-1em]
need a kiss  &  4.75 \\ \\[-1em]

\bottomrule
\end{tabular}
}
\caption{A sample of annotated tweets in English}
\label{data_sample}
\end{center}
\end{table*}

\subsection{Annotation}
We recruited annotators from Prolific.co and paid them \$15 USD per hour for their annotations. We set a ``first language'' requirement during annotator pre-screening. For example, an annotator must meet the requirement of ``Spanish as the first language'' to annotate the intimacy of Spanish tweets. Intimacy is annotated using a 5-point Likert scale where 1 indicates ``Not intimate at all'' and 5 indicates ``Very intimate''. In pilot annotations, we explored 7-point likert scales as well as Best-Worst-Scaling (BWS) similar to \citet{pei2020quantifying} and calculated the Krippendorff's $\alpha$ to measure inter-annotator agreement (IAA). We found that the 5-point-likert scale annotations ($\alpha$ = 0.38) achieve IAA similar to BWS ($\alpha$ = 0.36) and have higher IAA than a 7-point likert scale ($\alpha$ = 0.25). For each language in the training set, we collected annotations for 2,000 tweets. Because they annotated the intimacy of tweets, the annotators could see sexual or potentially offensive content during annotation. Therefore, we required annotators to be at least 18 years old to work on our task. Each annotator was explicitly notified about the potential for sexual or offensive content and they signed a consent form before starting the annotations.
Each tweet was annotated by 7 annotators and each annotator was shown 50 tweets. After the annotation, each annotator was required to complete a post-study survey about their demographics including gender, age, religion, country, education background, and occupation. For tweets that were not in the target language or that did not make sense (e.g. random characters), the annotators were instructed to annotate them as \emph{Invalid Tweet}. We use POTATO \citep{pei2022potato} to set up all the annotation interfaces.

\subsection{Quality control}
Annotating textual intimacy is challenging because of the subjective nature of intimacy perception and potential individual rating bias. We designed a series of quality control procedures throughout the annotation process: (1) we conducted 10 pilot studies on Prolific.com and revised our annotation procedures according to attained IAA and participant feedback; (2) annotation guidelines for each language were carefully translated by native translators,\footnote{Chinese, Spanish, Dutch, French, Korean, Portuguese, Hindi, and Italian guidelines are translated by native speakers in Snap Inc.~and the University of Michigan. The Arabic guideline was translated by one expert translator and one expert proofreader recruited from Upwork.com, and both translators were paid \$13/h.} which prompts the annotators to think about intimacy in their own languages; (3) all instructions in the recruitment phase were written in the annotator's indicated first language in the recruitment phase, which could potentially remove potential spam annotators in crowdsourcing platforms; (4) we randomly inserted two attention test questions\footnote{``This is a test question, please select N'' where N was a random number between 1-5} to identify potential spammers; (5) the annotators were balanced by sex (based on Prolific's built-in feature) and were also generally diverse regarding other demographics (e.g. the annotators are from 73 unique countries and regions), which allowed us to collect more population representative ratings.

\subsection{Post processing}
We first removed annotations from users who failed the attention test since they may have been spammers. No more than 2 annotators were removed in this step, except for Hindi (26 removed), Korean (7 removed), and Arabic (4 removed).
To remove potential noise in the crowdsourcing setting, similar to trimmed mean \citep{millsap2009sage}, we removed one highest score and one lowest score for tweets with at least five labels.  
After all the processing, we kept the tweets with at least two valid scores. For external test languages (i.e. Dutch, Hindi, Korean, and Arabic), we only kept tweets with a relatively low label diversity (i.e. standard deviation lower than 1) to ensure a good golden test set for the zero-shot setting\footnote{40\%, 15\%, 17\%, 16\% of tweets are removed for Hindi, Dutch, Korean, and Arabic respectively}. The final intimacy score is calculated as the mean score of all the remaining labels for each tweet.

\subsection{Annotation result}
\begin{figure}[t]
\centering
\includegraphics[width=0.45\textwidth]{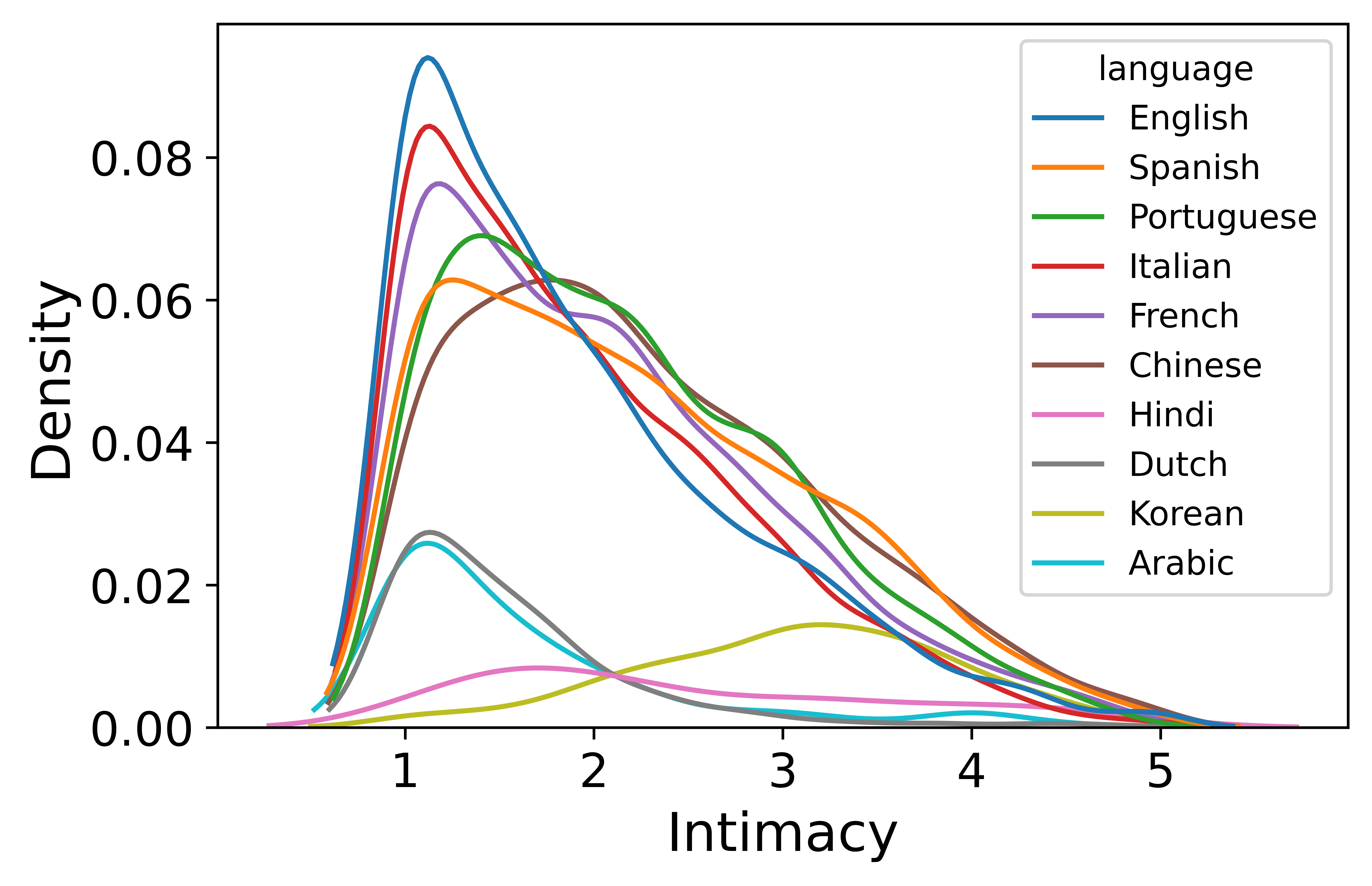}
\caption{The distribution of intimacy scores for each language}
%
\label{fig:intimacy-distribution}
\end{figure}

The final dataset includes 13,372 tweets annotated with the textual intimacy score. Table \ref{tab:dataset_stats} shows the final statistics for the annotated data. We attained moderate inter-annotator-agreement, similar to previous work \citep{pei2020quantifying}. Given the subjective nature of intimacy perception, we believe that such an IAA score is promising. To further verify the quality of the annotations, we conduct a split-half-reliability test \citep[SHR;][]{Johnson2022SplitHalfR}: randomly splitting labels into two groups and calculating the Pearson correlation between the aggregated scores from the two groups. All the SHR scores are above 0.63 with an average of 0.68, suggesting that the final aggregated scores are reliable. Figure \ref{fig:intimacy-distribution} shows the intimacy distribution of the final dataset. 
The final dataset for English, Spanish, French, Italian, Portuguese, and Chinese is split into training, validation, and test sets following a ratio of 7:1:2, and all the tweets are held as the test set for Arabic, Dutch, Korean, and Hindi.

\begin{table}[t]
\centering
\begin{adjustbox}{max width=\textwidth}
\begin{tabular}{cccc}
\toprule
\textbf{Language} & \textbf{$\alpha$} & \textbf{SHR} & \textbf{Amount} \\
\midrule
English & 0.48 & 0.69 & 1,983\\ 
Spanish & 0.52 & 0.72 & 1,991\\
Portuguese & 0.45 & 0.66 & 1,994 \\
Italian & 0.43 & 0.63 & 1,916\\
French & 0.47 & 0.67 & 1,981 \\
Chinese & 0.44  & 0.64 & 1,996 \\ \midrule
Hindi & 0.61 & 0.68 & \hspace{0.5em} 280 \\
Korean & 0.53 & 0.67 & \hspace{0.5em} 411 \\
Dutch & 0.48 & 0.68 & \hspace{0.5em} 413 \\ 
Arabic & 0.58 & 0.74 & \hspace{0.5em} 407 \\ \midrule
\bottomrule

\end{tabular}
\end{adjustbox}
\caption{Statistics for the annotated dataset}
\label{tab:dataset_stats}
\end{table}

\section{Baseline Models}
We benchmark several baseline models on the tweet intimacy prediction task. We compare the following multilingual pre-trained language models:
\begin{enumerate}
    \item BERT \citep{devlin2018bert}: multilingual BERT model.
    \item XLM-R \citep{conneau2019unsupervised}: multilingual RoBERTa model.
    \item XLM-T \citep{barbieri2021xlm}: Multilingual RoBERTa model trained over 200M tweets. 
    \item DistillBERT \citep{Sanh2019DistilBERTAD}: Multilingual distilled BERT model. 
    \item MiniLM \citep{wang2020minilm}: Multilingual MiniLM model. 
\end{enumerate}

All the models are trained with 10 epochs and the best performing model is selected based on the dev set\footnote{we evaluate the model performance every 500 steps and choose the best model}. We train each model with 5 different random seeds and report the mean score. The learning rate is set as 0.001 and the batch size is 64. We use AdamW as the optimizer. 

\begin{table}[t]
\resizebox{0.49\textwidth}{!}{
\begin{tabular}{cccccc}
\toprule
model &  XLM-T &  BERT &  XLM-R &  DistillBERT &  MiniLM \\
\midrule
English &      \textbf{0.70} &        0.59 &     0.65 &     0.55 &         0.61 \\
Spanish &      \textbf{0.73} &        0.62 &     0.64 &     0.61 &         0.67 \\
Portuguese       &      \textbf{0.65} &        0.54 &     0.61 &     0.52 &         0.53 \\
Italian &     \textbf{0.70} &        0.57 &     0.67 &     0.58 &         0.62 \\
French  &     \textbf{ 0.68} &        0.55 &     0.63 &     0.54 &         0.57 \\
Chinese &      0.70 &        0.65 &     \textbf{0.72} &     0.67 &         0.65 \\ \midrule
Hindi   &      \textbf{0.24} &        0.09 &     0.24 &     0.17 &         0.18 \\
Dutch   &      0.59 &        0.47 &     \textbf{0.60} &     0.44 &         0.57 \\
Korean  &      0.35 &        0.32 &     0.33 &     0.26 &         \textbf{0.41} \\
Arabic  &      \textbf{0.64} &        0.35 &     0.48 &     0.32 &         0.38 \\ \midrule
overall &      \textbf{0.58} &        0.48 &     0.53 &     0.52 &         0.53 \\
\bottomrule
\end{tabular}
}
\caption{Performance of the baselines. The bottom four rows are tested under the zero-shot setting.}
\label{tab:baseline_perf}
\end{table}

Table \ref{tab:baseline_perf} shows the performance of the baselines. We found that XLM-T achieved the best performance over 7 languages, suggesting that domain-specific language model training is beneficial for our tweet intimacy analysis task. 

For zero-shot tasks, while XLM-T still performs the best on Hindi and Arabic, XLM-R and MiniLM achieved the best result on Dutch and Korean, respectively. Moreover, the zero-shot performance is generally lower compared with the tasks with in-domain training, suggesting that the zero-shot task is challenging. We encourage the Task participants to explore different strategies to improve the zero-shot intimacy prediction performance.

\section{Conclusion}
In this paper, we present \DATASET{}, the dataset for \textit{SemEval 2023 Task 9: Multilingual Tweet Intimacy Analysis} and benchmarked a series of multilingual pre-trained language models. While the models are generally able to predict textual intimacy with promising performance, how to infer textual intimacy for unseen language remains challenging especially for Korean and Hindi. We encourage the SemEval participants to explore advanced methods for both monolingual and multilingual tweet intimacy analysis tasks.

\section{Acknowledgment} Our multilingual tweet intimacy annotation has been approved by the IRB office of the University of Michigan (HUM00214259).

\bibliography{anthology,custom}
\bibliographystyle{acl_natbib}




\end{document}